\def\year{2022}\relax
\documentclass[letterpaper]{article} 
\usepackage{aaai22}  
\usepackage{times}  
\usepackage{helvet}  
\usepackage{courier}  
\usepackage[hyphens]{url}  
\usepackage{graphicx} 
\usepackage{natbib}  
\usepackage{caption} 
\DeclareCaptionStyle{ruled}{labelfont=normalfont,labelsep=colon,strut=off} 
\usepackage[noend]{algpseudocode}
\usepackage[ruled,linesnumbered]{algorithm2e}
\usepackage{amsmath}

\usepackage{times}
\usepackage{url}
\usepackage{pifont}
\usepackage{multirow}
\usepackage{appendix}
\usepackage{mathtools}
\usepackage{booktabs}
\usepackage{amsthm}
\usepackage{tikz}
\usepackage{xcolor}

\DeclareMathOperator*{\argmin}{argmin}

\title{Parameter Differentiation based Multilingual Neural Machine Translation}
\author{Qian Wang$^{1,2}$
, Jiajun Zhang$^{1,2}$\thanks{Corresponding author: Jiajun Zhang.} \\ 
$^{1}$National Laboratory of Pattern Recognition, CASIA, Beijing, China \\
$^{2}$University of Chinese Academy of Sciences, Beijing, China \\
\{qian.wang, jjzhang\}@nlpr.ia.ac.cn}

\begin{document}

\maketitle

\begin{abstract}

Multilingual neural machine translation (MNMT) aims to translate multiple languages with a single model and has been proved successful thanks to effective knowledge transfer among different languages with shared parameters.
However, it is still an open question which parameters should be shared and which ones need to be task-specific.
Currently, the common practice is to heuristically design or search language-specific modules, which is difficult to find the optimal configuration.
In this paper, we propose a novel parameter differentiation based method 
that allows the model to determine which parameters should be language-specific during training.
Inspired by cellular differentiation,
each shared parameter in our method can dynamically differentiate into more specialized types.
We further define the differentiation criterion as inter-task gradient similarity.
Therefore, parameters with conflicting inter-task gradients are more likely to be language-specific. 
Extensive experiments on multilingual datasets
have demonstrated that our method significantly outperforms 
various strong baselines with different 
parameter sharing configurations.
Further analyses reveal that the parameter sharing configuration obtained by our method correlates well with the linguistic proximities.

\end{abstract}

\section{Introduction}

Neural machine translation (NMT) has achieved great success and drawn much attention in recent years
\cite{sutskever2014sequence,bahdanau2014neural,vaswani2017attention}.
While conventional NMT can well handle the translation of a single language pair, 
training an individual model for each language pair is resource-consuming, 
considering there are thousands of languages in the world. 
Therefore, multilingual NMT is developed to handle multiple language pairs in one model, 
greatly reducing the cost of offline training and online deployment \cite{ha2016toward2,johnson2017google}.
Besides, 
the parameter sharing in multilingual neural machine translation 
encourages positive knowledge transfer among different languages
and benefits low-resource  translation
\cite{DBLP:conf/acl/ZhangWTS20opus,siddhantJTARBFR20evalmnmt}.


\definecolor{shared}{HTML}{FFD9C3}
\definecolor{spec1}{HTML}{ABC2FF}
\definecolor{spec2}{HTML}{9AEA8B}
\begin{figure}[t]
\centering
\includegraphics[width=8cm]{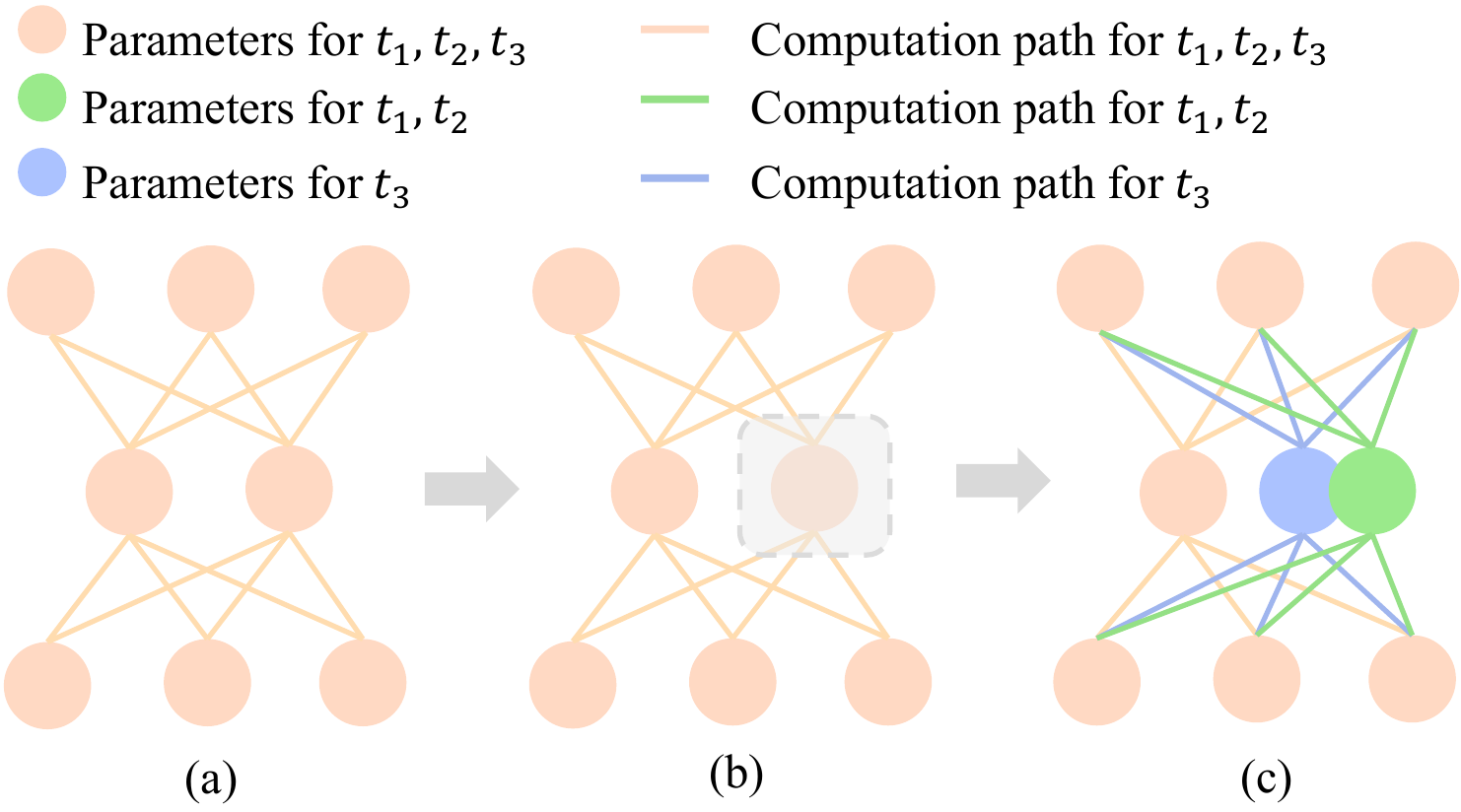}
\caption{
The illustration of parameter differentiation.
Each task $t_i$ represents a translation direction, e.g. EN$\rightarrow$DE.
(a) Initialized as completely shared, 
(b) the model detects parameters that should be more specialized during training,
and (c) the shared parameters differentiated
(\tikz[baseline=-0.5ex]\draw[shared,fill=shared] circle (1ex);) 
into more specialized types (\tikz[baseline=-0.5ex]\draw[spec1,fill=spec1] circle (1ex); 
\tikz[baseline=-0.5ex]\draw[spec2,fill=spec2] circle (1ex);).
\label{fig:arch}}
\end{figure}



Despite the benefits of the joint training with a completely shared model,
the MNMT model also suffers from insufficient model capacity \cite{arivazhagan2019massivelywild,DBLP:conf/emnlp/LyuSYB20mnmtindus}.
The shared parameters tend to preserve the general knowledge
but ignore language-specific knowledge.
Therefore, researchers resort to heuristically design additional language-specific components
and build MNMT model with a mix of shared and language-specific parameters to increase the model capacity \cite{DBLP:conf/wmt/SachanN18,wang2019compact},
such as the language-specific attention \cite{DBLP:conf/coling/BlackwoodBW18taskattn},
lightweight language adapter \cite{DBLP:conf/emnlp/BapnaF19simple}
or language-specific routing layer \cite{DBLP:conf/iclr/ZhangBSF21sharenot}.
These methods simultaneously model the general knowledge and the language-specific knowledge
but require specialized manual design.
Another line of works for language-specific modeling aims to automatically search for 
language-specific sub-networks
\cite{DBLP:conf/acl/Xie0G020,DBLP:conf/acl/LinWWL20langsubnet},
in which they pretrain an initial large model that covers all translation directions,
followed by sub-network pruning and fine-tuning.
These methods include multi-stage training and it is non-trivial to determine the initial model size and structure.


In this study,
we propose a novel parameter differentiation based method that enables 
the model to automatically determine which parameters should be shared and which ones should be language-specific during training.
Inspired by cellular differentiation, 
a process in which a cell changes from one general cell type to a more specialized type,
our method allows each parameter that shared by multiple tasks to dynamically differentiate into more specialized types.
As shown in Figure \ref{fig:arch},
the model is initialized as completely shared and continuously detects shared parameters that should be language-specific.
These parameters are then duplicated and reallocated to different tasks to increase language-specific modeling capacity.
The differentiation criterion is defined as inter-task gradient similarity,
which represents the consistency of optimization direction across tasks on a shared parameter.
Therefore,
the parameters facing conflicting inter-task gradients are selected for differentiation
while other parameters with more similar inter-task gradients remain shared.
In general, 
the MNMT model in our method can gradually improve its parameter sharing configuration 
without multi-stage training or manually designed language-specific modules.

We conduct extensive experiments on three widely used multilingual datasets including OPUS, WMT and IWSLT
in multiple MNMT scenarios: one-to-many, many-to-one and many-to-many translation.
The experimental results prove the effectiveness of the proposed method over various strong baselines.
Our main contributions can be summarized as follows:

\begin{quote}
\begin{itemize}
\item We propose a method that can automatically determine which parameters in an MNMT model should be language-specific without manual design,
and can dynamically change shared parameters into more specialized types.
\item We define the differentiation criterion as the inter-task gradient similarity, which helps to minimizes the inter-task interference on shared parameters.
\item We show that the parameter sharing configuration obtained by our method is highly correlated with linguistic features like language families.
\end{itemize}
\end{quote}

\section{Background}
\paragraph{The Transformer Model}
A typical Transformer model  \cite{vaswani2017attention}
consists of an encoder and a decoder.
Both the encoder and the decoder are stacked with $N$ identical layers.
Each encoder layer contains two modules named 
multi-head self-attention and feed-forward network.
The decoder layer, containing three modules,
inserts an additional multi-head cross-attention between the self-attention and feed-forward modules.

\paragraph{Multilingual Neural Machine Translation}
The standard paradigm of MNMT contains a completely shared model borrowed from bilingual translation for all language pairs.
A special language token is appended to the source text to indicate the target language,
i.e., $X=\{\text{lang}, x_1, \dots, x_n \}$
\cite{johnson2017google}.
The MNMT is often referred to as multi-task optimization,
in which a \textit{task} indicates a translation direction, e.g. EN$\rightarrow$DE.


\section{Parameter Differentiation based MNMT}

Our main idea is to find out shared parameters 
that should be language-specific in an MNMT model
and dynamically change them into more specialized types during training.
To achieve this, we propose a novel parameter differentiation based MNMT approach 
and define the differentiation criterion as inter-task gradient similarity.


\SetKwInOut{Input}{Input}
\SetKwInOut{Output}{Output}
\SetKw{Continue}{continue}
\DontPrintSemicolon

\begin{algorithm}[t]
\small
\caption{Parameter Differentiation \label{algo:pd}}
\Input{training data $\mathcal{D}$, Tasks $T=\{t_1, t_2, \dots\}$, \\
models for each task $\mathcal{M}=\{\mathcal{M}_{t_1}, \mathcal{M}_{t_2}, \dots\}$ }
\tcp{Initialize the shared model} 
$\mathcal{M}_{t_1}$ = $\mathcal{M}_{t_2}$ = $\mathcal{M}_{t_3}$ = \dots \;
\While{$\mathcal{M}$ not converge}{
    Train the model $\mathcal{M}$ with data $\mathcal{D}$ \;
    \tcp{Detect parameters to differentiate}
    flagged = [] \;
    \For{each $\theta_i$ in shared parameters of $\mathcal{M}$}{
        Evaluate $\theta_i$ with differentiation criterion \;
        \If{$\theta_i$ should be language-specific}{
            Add $\theta_i$ into flagged \;
        }
    }
    \tcp{Reallocate parameters}
    \For{each $\theta_i$ shared by tasks $T_i$ in flagged}{
        Split $T_i$ into $T_{i'}$ and $T_{i''}$ \;
        Duplicate $\theta_i$ into $\theta_{i'}$, $\theta_{i''}$\;
        Replace $\theta_{i}$ in $\mathcal{M}_{t| t \in T_{i'}}$ with $\theta_{i'}$ \;
        Replace $\theta_{i}$ in $\mathcal{M}_{t| t \in T_{i''}}$ with $\theta_{i''}$ \;
    }
}
\end{algorithm}

\subsection{Parameter Differentiation\label{sec:pd}}

As we know that cellular differentiation is the process in which a cell changes from one cell type to another, 
typically from a less specialized type (stem cell) to a more specialized type (organ/tissue-specific cell)
\cite{slack2007metaplasiacelldiff2}.
Inspired by cellular differentiation,
we propose parameter differentiation that can
dynamically change the task-agnostic parameters in an MNMT model into other task-specific types during training.





Algorithm \ref{algo:pd} lists the overall process of our method.
We first initialize the completely shared MNMT model following the paradigm in \cite{johnson2017google}.
After training for several steps, the model evaluates each shared parameter
and flag the parameters that should become more specialized under a certain \textbf{differentiation criterion}
(Line $4$-$10$).
For those flagged parameters, the model then duplicates them and reallocates the replicas for different tasks.
After the duplication and reallocation,
the model builds new connections for those replicas to construct different computation graphs 
$\mathcal{M}_{t_j}$ for each task (Line $11$-$16$).
In the following training steps,
the parameters belonging to $\mathcal{M}_{t_j}$ only update
for training data of task $t_j$.
The differentiation happens after every several training steps 
and the model dynamically becomes more specialized.

\definecolor{sharedtxt}{HTML}{C5A796}
\definecolor{spec1txt}{HTML}{78B76C}
\definecolor{spec2txt}{HTML}{8596C6}
\definecolor{optim}{HTML}{C00000}
\begin{figure}[t]
\vspace{5mm}
\centering
\includegraphics[width=8cm]{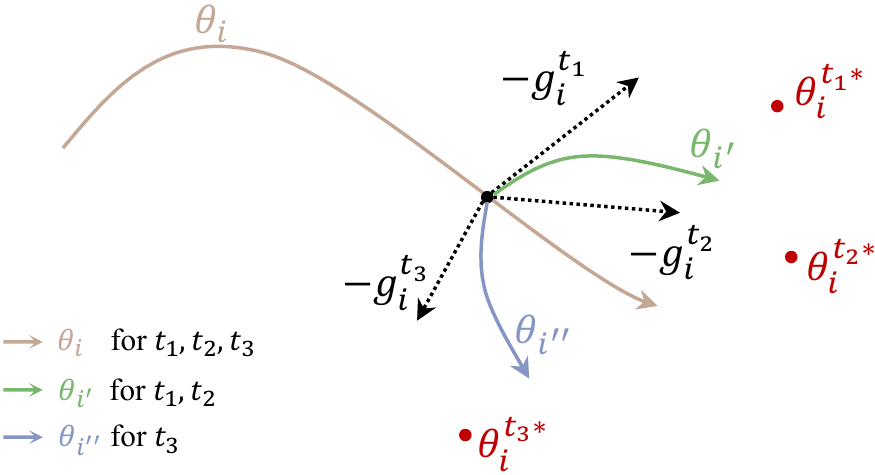}
\vspace{5mm}
\caption{The illustration of parameter differentiation with gradient cosine similarity.
The shared parameter \textcolor{sharedtxt}{$\theta_i$} differentiates into 
\textcolor{spec1txt}{$\theta_{i'}$} for tasks $\{t_1, t_2\}$
and \textcolor{spec2txt}{$\theta_{i''}$} for $\{ t_3\}$ respectively
since the gradients $g_i^{t_1}$ and $g_i^{t_2}$ are more similar.
\textcolor{optim}{$\theta_i^{t_j*}$} denotes the global optimum of 
$\theta_i$ on task $t_j$.
\label{fig:optim}}
\end{figure}

\subsection{The Differentiation Criterion\label{sec:crit}}

The key issue in parameter differentiation is the definition of differentiation criterion
that helps to detect the shared parameters that should differentiate into more specialized types.
We define the differentiation criterion based on inter-task gradient cosine similarity,
where the parameters facing conflicting gradients are more likely to be language-specific.

As shown in Figure \ref{fig:optim},
the parameter $\theta_i$ is shared by tasks $t_1$, $t_2$, and $t_3$ at the beginning.
To determine whether the shared parameter should be more specialized, 
we first define the \textbf{interference degree} of the parameter shared by the three tasks 
with the inter-task gradient cosine similarity.
More formally,
suppose the $i$-th parameter $\theta_i$ in an MNMT model is shared by a set of tasks $T_i$,
the interference degree $\mathcal{I}$ of the parameter $\theta_i$ is defined by:
\begin{equation} 
\begin{split}
     \mathcal{I}(\theta_i, T_i) = 
     \max_{t_j, t_k \in T_i} - \frac{g^{t_j}_i \cdot g^{t_k}_i  }{\| g^{t_j}_i \|  \| g^{t_k}_i \|}
\end{split}
\end{equation}
where $g^{t_j}_i$ and $g^{t_k}_i$ are the gradients of task $t_j$ and $t_k$ respectively
on the parameter $\theta_i$.

Intuitively,
the gradients determine the optimization directions.
For example in Figure \ref{fig:optim},
the gradient $g^{t_j}_i$ indicates the direction of global optimum for task $t_j$.  
The gradients with maximum negative cosine similarity,
such as $g^{t_1}_i$ and $g^{t_3}_i$,
point to opposite directions,
which hinders the optimization and has been proved detrimental for multi-task learning
\cite{DBLP:conf/nips/YuK0LHF20gradsug,DBLP:conf/iclr/WangTF021gradvacc}. 

The gradients of each task on each shared parameter are evaluated on \textbf{held-out validation data}.
To minimize the gradient variance caused by inconsistent sentence semantics across languages,
the validation data is created as multi-way aligned,
i.e., each sentence has translations of all languages.
With the held-out validation data,
we evaluate gradients of each task on each shared parameter 
for calculating inter-task gradient similarities as well as 
the interference degree $\mathcal{I}$ for each parameter.

The interference degree $\mathcal{I}$ helps the model to find out parameters that face severe interference and
the parameters with high interference degrees are flagged for differentiation.
Suppose the parameter $\theta_i$ shared by tasks $T_i$ is flagged,
we cluster the tasks in $T_i$ into two subsets $T_{i'}$ and $T_{i''}$ that minimize the overall interference.
The partition $P^*$ is obtained by:
\begin{equation}
\begin{split}
    P^*_i = \argmin_{T_{i'}, T_{i''}} [\mathcal{I}(\theta_i, T_{i'}) + \mathcal{I}(\theta_i, T_{i''})]
\end{split}
\end{equation}

As shown in Figure \ref{fig:optim},
the gradients of $g_i^{t_1}$ and $g_i^{t_2}$ are similar
while $g_i^{t_1}$ and $g_i^{t_3}$ are in conflict with each other.
By minimizing the overall interference degree,
the tasks are clustered into partition $P^*: T_{i'} = \{t_1, t_2\}, T_{i''} = \{t_3\} $.
The parameter $\theta_i$ is then duplicated into $\theta_{i'}$ and $\theta_{i''}$
and the replicas are allocated to $T_{i'}$ and $T_{i''}$ respectively.

\subsection{The Differentiation Granularity}

\begin{table}[t]
\small
\begin{center}
\begin{tabular}{cc} 
\specialrule{0em}{0.9pt}{0.9pt} \toprule[0.8pt] \specialrule{0em}{0.9pt}{0.9pt}
Granularity & Examples of Differentiation Units  \\ 
\specialrule{0em}{0.9pt}{0.9pt} \hline \specialrule{0em}{0.9pt}{0.9pt}
\multirow{1}{*}{\centering Layers}        
& encoder layer, decoder layer   \\ \specialrule{0em}{0.9pt}{0.9pt}
\multirow{1}{*}{\centering Module}  
& {self-attention}, {feed-forward}, {cross-attention}   \\ \specialrule{0em}{0.9pt}{0.9pt}
\multirow{1}{*}{\centering Operation}  
& linear projection, layer normalization   \\ 
\specialrule{0em}{0.9pt}{0.9pt} 
\bottomrule[0.8pt]
\end{tabular} 
\end{center}
\caption{The examples of differentiation units under different granularities.
\label{tab:granu}}
\end{table}

In theory, each shared parameter can differentiate into more specialized types individually.
But in practice,
performing differentiation on every single parameter is resource- and time-consuming,
considering there are millions to billions of parameters in an MNMT model.

Therefore, we resort to different levels of differentiation granularity,
like Layer, Module, or Operation.
As shown in Table \ref{tab:granu},
the Layer granularity indicates different layers in the model,
while the Module granularity specifies the individual modules within a layer.
The Operation granularity includes the basic transformations in the model that contain trainable parameters.
With a certain granularity, the parameters are grouped into different differentiation units.
For example,
with the Layer level granularity, the parameters within a layer are concatenated into a vector and differentiate together,
where the vector is referred to as a differentiation unit. 
We list the differentiation units in supplementary materials. 


\subsection{Training}

In our method, since the model architecture dynamically changes 
and results in different computation graphs for each task,
we create batches from the multilingual dataset and ensure that each batch contains only samples from one task.
This is different from the training of vanilla completely shared MNMT model
where each batch may contain sentence pairs from different languages \cite{johnson2017google}.
Specifically, 
we first sample a task $t_j$,
followed by sampling a batch $\mathcal{B}_{t_j}$ from training data of $t_j$.
Then, the model $\mathcal{M}_{t_j}$
which includes a mix of shared and language-specific parameters
is trained with the batch $\mathcal{B}_{t_j}$.


We train the model with the Adam optimizer \cite{DBLP:journals/corr/KingmaB14},
which computes adaptive learning rates based on the optimizing trajectory of past steps.
However, the optimization history becomes inaccurate for the differentiated parameters.
For the example in Figure \ref{fig:optim},
the differentiated parameter $\theta_{i''}$ is only shared by task $t_3$,
while the optimization history of $\theta_{i}$ represents the optimizing trajectory of all the $3$ tasks.
To stabilize the training of $\theta_{i''}$ on task $t_3$,
we reinitialize the optimizer states by performing a warm-up update for those differentiated parameters:
\begin{align} 
\begin{split} 
m_t' = \beta_1 m_{t} +& (1 - \beta_1) (g_i^{t_3}) \\ 
v_t' = \beta_2 v_{t} +& (1 - \beta_2) (g_i^{t_3})^2 
\end{split} 
\end{align}
where $m_{t}$ and $v_{t}$ are the Adam states of $\theta_i$, 
and $g_i^{t_3} $ is the gradient of task $t_3$ on the held-out validation data.
Note that we only update the states in the Adam optimizer and the parameters remain unchanged
in the warm-up update step.

\section{Experiments}

\subsection{Dataset}

We use the public OPUS and WMT multilingual datasets to evaluate our method 
on many-to-one (M2O) and one-to-many (O2M) translation scenarios,
and the IWSLT datasets for the many-to-many (M2M) translation scenario.

The OPUS dataset consists of English to $12$ languages
selected from the original OPUS-100 dataset
\cite{DBLP:conf/acl/ZhangWTS20opus}.
These languages, containing $1$M sentences for each,
are from $6$ distinct language groups:
Romance (French, Italian), 
Baltic (Latvian, Lithuanian),
Uralic (Estonian, Finnish),
Austronesian (Indonesian, Malay),
West-Slavic (Polish, Czech) and
East-Slavic (Ukrainian, Russian).


The WMT dataset with unbalanced data distribution is collected from the WMT'14, WMT'16 and WMT'18 benchmarks.
We select $5$ languages
with data sizes 
ranging from $0.6$M to $39$M.
The training data sizes and sources are shown in Table \ref{tab:tab1}.
We report the results on the WMT dataset with the temperature-based sampling 
in which the temperature is set to $\tau=5$ 
\cite{arivazhagan2019massivelywild}.

We evaluate our method on the many-to-many scenario with the IWSLT'17 dataset,
which includes German, English, Italian, Romanian, and Dutch,
and results in $20$ translation directions between the $5$ languages.
Each translation direction contains about $200$k sentence pairs.


The held-out multi-way aligned validation data for measuring gradient similarities
contains $4,000$ sentences for each language,
and are randomly selected and excluded from the training set.
We apply the byte-pair encoding (BPE) algorithm \cite{DBLP:conf/acl/SennrichHB16abpe}
with vocabulary sizes of $64$k for both OPUS and WMT datasets, and $32$k for the IWSLT dataset.

\begin{table}[!t]
\small
\begin{center}
\begin{tabular}{ccc} 
\specialrule{0em}{0.9pt}{0.9pt} \toprule[0.8pt] \specialrule{0em}{0.9pt}{0.9pt}
Language Pair    & Data Source & \#Samples \\ 
\specialrule{0em}{0.9pt}{0.9pt} \hline \specialrule{0em}{0.9pt}{0.9pt}
English-French (EN-FR) & WMT'14 & $39.03$M \\ \specialrule{0em}{0.9pt}{0.9pt}
English-Czech (EN-CS)  & WMT'14  & $15.65$M \\ \specialrule{0em}{0.9pt}{0.9pt}
English-German (EN-DE)  & WMT'14  & $4.46$M \\ \specialrule{0em}{0.9pt}{0.9pt}
English-Estonian (EN-ET)  & WMT'18  & $1.94$M \\ \specialrule{0em}{0.9pt}{0.9pt}
English-Romanian (EN-RO)  & WMT'16  & $0.61$M \\ 
\specialrule{0em}{0.9pt}{0.9pt} 
\bottomrule[0.8pt]
\end{tabular} 
\end{center}
\caption{Training data sizes and sources for the unbalanced WMT dataset.
\label{tab:tab1}}
\end{table}


\begin{table*}[t!] \small
\vspace*{1mm}
\begin{center}
\begin{tabular}{ccccccccccccccc} 
\specialrule{0em}{0.80pt}{0.80pt}\toprule[0.8pt] \specialrule{0em}{0.80pt}{0.80pt}
& Languages 
& \multicolumn{2}{c}{FR$\leftrightarrow$EN} & \multicolumn{2}{c}{IT$\leftrightarrow$EN} 
& \multicolumn{2}{c}{LV$\leftrightarrow$EN} 
& \multicolumn{2}{c}{LT$\leftrightarrow$EN} & \multicolumn{2}{c}{ET$\leftrightarrow$EN} \\
\specialrule{0em}{0.80pt}{0.80pt}
& Direction & $\leftarrow$  & $\rightarrow$  & $\leftarrow$  & $\rightarrow$ 
& $\leftarrow$  & $\rightarrow$ & $\leftarrow$  & $\rightarrow$ & $\leftarrow$  & $\rightarrow$ \\ 
\specialrule{0em}{0.80pt}{0.80pt}\hline \specialrule{0em}{0.80pt}{0.80pt}
\multirow{5}{*}{\centering Baselines} 
& Bilingual \cite{vaswani2017attention} & 
28.90 & 28.27 & 22.55 & 25.55 & 31.60 & 39.75 & 28.88 & 36.43 & 18.65 & 25.48 \\ 
\specialrule{0em}{0.80pt}{0.80pt}
& Multilingual \cite{johnson2017google} & 
27.33 & 28.31 & 21.20 & 27.09 & 30.00 & 40.10 & 27.69 & 37.15 & 20.08 & 30.09  \\
\specialrule{0em}{0.80pt}{0.80pt}
& Random Sharing
& 27.48 & 28.91 & 21.42 & 27.18 & 31.57 & 41.18 & 28.94 & 37.57 & 20.43 & 30.15 \\
\specialrule{0em}{0.80pt}{0.80pt}
& \citet{DBLP:conf/emnlp/TanCHXQL19cluster} & 
27.39 & 29.21 & 21.97 & 26.77 & 31.85 & \textbf{42.71} & 29.27 & 39.34 & {21.40} & 29.79  \\ 
\specialrule{0em}{0.80pt}{0.80pt}
&  \citet{DBLP:conf/wmt/SachanN18} & 
28.04 & 29.31 & 22.86 & 27.86 & 32.04 & 41.43 & 28.47 & 38.14 & \textbf{21.41} & 30.30 \\ 
\specialrule{0em}{0.80pt}{0.80pt}\hline \specialrule{0em}{0.80pt}{0.80pt}
\multirow{3}{*}{\centering Ours} 
& PD w. Layer & 
\textbf{29.35} & 30.09 & 22.37 & \textbf{28.7} & 32.31 & 42.11 & 29.5 & 39.04 & 20.56 & 30.91 \\
\specialrule{0em}{0.80pt}{0.80pt}
& PD w. Module & 
29.09 & 30.09 & 22.49 & 28.64 & 31.86 & 41.60 & 29.53 & 39.04 & 21.25 & 31.11 \\
\specialrule{0em}{0.80pt}{0.80pt}
& PD w. Operation & 
29.26 & \textbf{30.11} & \textbf{23.01} & 28.6 & \textbf{33.06} & {42.38} 
& \textbf{29.94} & \textbf{39.54} & 20.89 & \textbf{31.14}  \\
\specialrule{0em}{0.80pt}{0.80pt}
\hline \specialrule{0em}{0.5pt}{0.5pt}  \hline
\specialrule{0em}{0.80pt}{0.80pt}
& Languages
& \multicolumn{2}{c}{FI$\leftrightarrow$EN} & \multicolumn{2}{c}{ID$\leftrightarrow$EN}
& \multicolumn{2}{c}{MS$\leftrightarrow$EN} & \multicolumn{2}{c}{PL$\leftrightarrow$EN}
& \multicolumn{2}{c}{CS$\leftrightarrow$EN}  \\ 
\specialrule{0em}{0.80pt}{0.80pt}
& Direction & $\leftarrow$  & $\rightarrow$  & $\leftarrow$  & $\rightarrow$ 
& $\leftarrow$  & $\rightarrow$ & $\leftarrow$  & $\rightarrow$ & $\leftarrow$  & $\rightarrow$ \\ 
\specialrule{0pt}{1pt}{1pt} \hline \specialrule{0em}{0.80pt}{0.80pt}
\multirow{5}{*}{\centering Baselines} 
& Bilingual \cite{vaswani2017attention} & 
13.92 & 18.34 & 21.29 & 25.61 & 16.75 & 21.24 & 13.46 & 19.05 & 16.82 & 25.27 \\ 
\specialrule{0em}{0.80pt}{0.80pt}
& Multilingual \cite{johnson2017google} & 
15.58 & 21.43 & 22.85 & 28.27 & 18.12 & 23.66 & 14.87 & 22.24 & 18.57 & 28.14  \\ 
\specialrule{0em}{0.80pt}{0.80pt}
& Random Sharing
& 16.01 & 21.30 & 21.69 & 27.78 & 17.13 & 23.73 & 15.23 & 21.97 & 18.40 & 28.21  \\
\specialrule{0em}{0.80pt}{0.80pt}
& \citet{DBLP:conf/emnlp/TanCHXQL19cluster} & 
16.15 & 21.46 & 22.74 & 28.00 & 18.12 & 23.14 & 14.86 & 21.72 & 18.02 & 28.08 \\ 
\specialrule{0em}{0.80pt}{0.80pt}
&  \citet{DBLP:conf/wmt/SachanN18} & 
16.37 & 21.36 & 22.39 & \textbf{29.60} & 17.33 & 23.77 & 15.75 & 22.45 & \textbf{19.70} & 28.59 \\ 
\specialrule{0em}{0.80pt}{0.80pt}\hline \specialrule{0em}{0.80pt}{0.80pt}
\multirow{3}{*}{\centering Ours} 
& PD w. Layer & 
16.42 & 22.37 & 22.89 & {29.28} & 18.35 & 24.88 & 16.07 & 23.11 & 19.29 & 29.31 \\ 
\specialrule{0em}{0.80pt}{0.80pt}
& PD w. Module & 
16.44 & \textbf{22.85} & \textbf{22.94} & 28.86 & 17.62 & 24.27 & 16.18 & 23.12 & 19.33 & 29.08 \\ 
\specialrule{0em}{0.80pt}{0.80pt}
& PD w. Operation & 
\textbf{16.59} & \textbf{22.85} & 23.09 & 29.03 & \textbf{18.61} & \textbf{25.27} 
& \textbf{16.45} & \textbf{23.34} & {19.46} & \textbf{29.66}  \\
\specialrule{0em}{0.80pt}{0.80pt}
\hline \specialrule{0em}{0.5pt}{0.5pt}  \hline
\specialrule{0em}{0.80pt}{0.80pt}
& Languages
& \multicolumn{2}{c}{UK$\leftrightarrow$EN} & \multicolumn{2}{c}{RU$\leftrightarrow$EN} 
& \multicolumn{2}{c}{Average} & \multicolumn{2}{c}{$\Delta$ Average} & \multicolumn{2}{c}{Model Size} \\ 
\specialrule{0em}{0.80pt}{0.80pt}
& Direction & $\leftarrow$  & $\rightarrow$  & $\leftarrow$  & $\rightarrow$ 
& $\leftarrow$  & $\rightarrow$ & $\leftarrow$  & $\rightarrow$ & $\leftarrow$  & $\rightarrow$ \\ 
\specialrule{0pt}{1pt}{1pt} \hline \specialrule{0em}{0.80pt}{0.80pt}
\multirow{5}{*}{\centering Baselines} 
& Bilingual \cite{vaswani2017attention} & 
10.06 & 18.68 & 21.63 & 26.61 & 20.38 & 25.86 & -0.36 & -2.22 & 12x & 12x \\ 
\specialrule{0em}{0.80pt}{0.80pt}
& Multilingual \cite{johnson2017google} & 
11.59 & 21.76 & 20.96 & 28.76 & 20.74 & 28.08 & 0 & 0 & 1x & 1x \\ 
\specialrule{0em}{0.80pt}{0.80pt}
& Random Sharing
& 11.57 & 21.83 & 21.36 & 28.91 & 20.93 & 28.23 & +0.19 & +0.15 & 1.98x & 2.00x \\
\specialrule{0em}{0.80pt}{0.80pt}
&  \citet{DBLP:conf/emnlp/TanCHXQL19cluster} & 
11.32 & 21.74 & 21.32 & 28.73 & 21.20 & 28.39 & +0.46 & +0.31 & 2x & 2x \\ 
\specialrule{0em}{0.80pt}{0.80pt}
&  \citet{DBLP:conf/wmt/SachanN18} &  10.96 & 21.88 & 22.28 & 28.80 & 21.47 & 28.62 & +0.73 & +0.54 &
3.71x & 3.25x \\ 
\specialrule{0em}{0.80pt}{0.80pt}\hline \specialrule{0em}{0.80pt}{0.80pt}
\multirow{3}{*}{\centering Ours} 
& PD w. Layer & 
12.32 & 22.68 & 22.82 & 30.37 & 21.85 & 29.40 & +1.11 & +1.32 & 2.14x & 1.84x \\ 
\specialrule{0em}{0.80pt}{0.80pt}
& PD w. Module & 
\textbf{12.55} & 22.44 & 22.31 & 30.39 & 21.80 & 29.29 & +1.06 & +1.21 & 1.82x & 1.94x \\ 
\specialrule{0em}{0.80pt}{0.80pt}
& PD w. Operation & 
12.37 & \textbf{23.05} & \textbf{22.98} & \textbf{30.60} & \textbf{22.14} & \textbf{29.63} &
+1.40 & +1.55 & 1.96x & 1.90x \\ 
\specialrule{0em}{0.80pt}{0.80pt}
\bottomrule[0.8pt]
\end{tabular}
\vspace{-1mm}
\end{center}
\caption{\label{tab:opus}
BLEU scores on the OPUS dataset.
We compare our method with different levels of parameter sharing in both
one-to-many ($\leftarrow$) and many-to-one ($\rightarrow$) directions. 
We report our parameter differentiation (PD) method with different granularity: 
Layer, Module and Operation.
\textbf{Bold} indicates the best result of all methods.}
\end{table*}

\subsection{Model Settings}
We conduct our experiments with the Transformer architecture
and adopt the \textit{transformer\_base} setting which includes
$6$ encoder and decoder layers, $512$/$2048$ hidden dimensions and $8$ attention heads.
Dropout ($p=0.1$) and label smoothing ($\epsilon_{ls}=0.1$)
are applied during training but disabled during validation and inference.
Each mini-batch contains roughly $8,192$ tokens.
We accumulate gradients and update the model every $4$ steps for OPUS and $8$ steps for WMT
to simulate multi-GPU training.
In inference, we use beam search with the beam size of $4$ 
and the length penalty of $0.6$.
We measure the translation quality by BLEU score \cite{papineni2002bleu}
with SacreBLEU\footnote{https://github.com/mjpost/sacrebleu}.
All the models are trained and tested on a single Nvidia V100 GPU.

Our method allows the parameters to differentiate into specialized types by duplication and reallocation,
which may results in bilingual models with unlimited parameter differentiation,
i.e., each parameter is only shared by one task in the final model.
To prevent over-specialization and make a fair comparison,
we set a differentiation upper bound defined by the expected final model size $\mathcal{O}$,
and let the model control the number of parameters (denoted as $k$) to differentiate\footnote{
Since the parameters are grouped into differentiation units under a certain granularity, 
the value of $k$ and $\mathcal{O}$ may fluctuate to comply with the granularity.}:
\begin{equation} 
\label{eq:eqmodel}
\begin{split}
              \mathcal{O} &\approx \mathcal{O}_0 + \frac{Q}{N} \times k \\
 \Rightarrow   k &\approx \frac{N}{Q} \times (\mathcal{O} - \mathcal{O}_0)
\end{split}
\end{equation}
where $\mathcal{O}_0$ is the size of the original completely shared model.
The total training step $Q$ is set to $400$k for all experiments,
and the differentiation happens every $N=8000$ steps of training.
We set the expected model size to ${\mathcal{O}} = 2\times\mathcal{O}_0$, $2$ times of original model.
We also analyze the relationship between model size and translation quality
by varying ${\mathcal{O}}$ in the range from $1.5$ to $4$.

\subsection{Baseline Systems}

We compare our method with several baseline methods 
with different paradigms of parameter sharing.

\textbf{Bilingual} trains Transformer model \cite{vaswani2017attention} for each translation direction
and results in $N$ individual models for $N$ translation directions.

\textbf{Multilingual} adopts the standard paradigm of MNMT
that all parameters are shared across tasks \cite{johnson2017google}.

\textbf{Random Sharing}
selects parameters for differentiation randomly (with Operation granularity)
instead of using inter-task gradient similarity.

\textbf{\citet{DBLP:conf/wmt/SachanN18}} 
uses a partially shared model that proved effective empirically. 
They share the attention key and query of the decoder, the embedding, and the encoder in a one-to-many model.
We extend the settings for the many-to-one model that share
the attention key and query of the encoder, the embedding, and the decoder.

\textbf{\citet{DBLP:conf/emnlp/TanCHXQL19cluster}}
first clusters the languages using the language embedding vectors in the \textbf{Multilingual} method
and then trains one model for each cluster.
To make the model size comparable with our method,
we set the number of clusters as $2$ and train two distinct models.
In our experiment on the OPUS dataset, 
this method results in two clusters:
$\{$FR, IT, ID, MS, PL, CS, UK, RU$\}$
and $\{$LV, LT, ET, FI$\}$.

\begin{table*}[t] \small
\begin{center}
\begin{tabular}{ccccccccccccccccc} 
\specialrule{0em}{0.85pt}{0.85pt} \toprule[0.8pt] \specialrule{0em}{0.85pt}{0.85pt}
Languages
& \multicolumn{2}{c}{FR$\leftrightarrow$EN} 
& \multicolumn{2}{c}{CS$\leftrightarrow$EN} 
& \multicolumn{2}{c}{DE$\leftrightarrow$EN} 
& \multicolumn{2}{c}{ET$\leftrightarrow$EN} 
& \multicolumn{2}{c}{RO$\leftrightarrow$EN} 
& \multicolumn{2}{c}{Average} 
& \multicolumn{2}{c}{Sizes} \\ 
\specialrule{0em}{0.85pt}{0.85pt}  \specialrule{0em}{0.85pt}{0.85pt}
Direction & $\leftarrow$  & $\rightarrow$  & $\leftarrow$  & $\rightarrow$ 
& $\leftarrow$  & $\rightarrow$ & $\leftarrow$  & $\rightarrow$ & $\leftarrow$  & $\rightarrow$ 
& $\leftarrow$  & $\rightarrow$ & $\leftarrow$  & $\rightarrow$ \\ 
\specialrule{0em}{0.85pt}{0.85pt} \hline \specialrule{0em}{0.85pt}{0.85pt}
Bilingual & 
39.87 & \underline{37.74} & \underline{27.23} & 31.43 & 26.71 & 31.98 & 17.55 & 23.26 & 23.13 & 29.23 & 26.90 & 30.73
& 5x & 5x \\ 
\specialrule{0em}{0.85pt}{0.85pt}
Multilingual & 
38.07 & 36.23 & 25.39 & 30.77 & 24.67 & 31.54 & 18.90 & 26.09 & 26.42 & 34.85 & 26.69 & 31.90 & 1x & 1x\\
\specialrule{0em}{0.85pt}{0.85pt} 
Ours &  
\textbf{\underline{40.28}} & \textbf{37.36} & \textbf{26.75} & \textbf{\underline{32.92}} &
\textbf{\underline{27.29}} & \textbf{\underline{32.80}} & \textbf{\underline{19.66}} & \textbf{\underline{27.64}} & 
\textbf{\underline{27.34}} & \textbf{\underline{35.90}} & \textbf{\underline{28.26}} & \textbf{\underline{33.32}} & 
1.83x & 1.87x \\
\bottomrule[0.8pt]
\end{tabular}
\end{center}
\caption{\label{tab:wmt}
Results on the WMT dataset.
Our method is parameter differentiation with granularity of Operation. 
\textbf{Bold} indicates the best result for multilingual model
while the overall best results are \underline{underlined}.}
\end{table*}

\begin{table*}[t]
\small
\begin{center}
\begin{tabular}{ccccccc} 
\specialrule{0em}{0.85pt}{0.85pt} \toprule[0.8pt] \specialrule{0em}{0.85pt}{0.85pt}
& $\rightarrow$DE & $\rightarrow$EN & $\rightarrow$IT & $\rightarrow$NL & $\rightarrow$RO & Average \\ 
\specialrule{0em}{0.85pt}{0.85pt} \hline \specialrule{0em}{0.85pt}{0.85pt}
DE$\rightarrow$  
& - & 33.09 / \textbf{34.07} & 20.10 / \textbf{21.05} & 22.18 / \textbf{22.35} & 18.20 / \textbf{19.00} & 23.39 / \textbf{24.12} \\ \specialrule{0em}{0.85pt}{0.85pt}
EN$\rightarrow$  & 26.55 / \textbf{27.86} & - & 27.74 / \textbf{28.69} & 28.15 / \textbf{28.61} & 25.34 / \textbf{26.58} & 26.95 / \textbf{27.94} \\ \specialrule{0em}{0.85pt}{0.85pt}
IT$\rightarrow$  & 19.32 / \textbf{20.37} & 32.14 / \textbf{32.99} & - & 20.05 / \textbf{20.47} & 19.28 / \textbf{20.10} & 22.70 / \textbf{23.48} \\ \specialrule{0em}{0.85pt}{0.85pt} 
NL$\rightarrow$  & 21.06 / \textbf{22.09} & 32.54 / \textbf{33.45} & 19.81 / \textbf{20.64} & - & 17.93 / \textbf{18.81} & 22.84 / \textbf{23.75} \\ \specialrule{0em}{0.85pt}{0.85pt} 
RO$\rightarrow$  & 20.74 / \textbf{21.39} & 34.78 / \textbf{35.76} & 22.96 / \textbf{23.53} & 20.87 / \textbf{21.04} & - & 24.84 / \textbf{25.43} \\ \specialrule{0em}{0.85pt}{0.85pt} 
Average & 21.92 / \textbf{22.93} & 33.14 / \textbf{34.07} & 22.65 / \textbf{23.48} & 22.81 / \textbf{23.12} & 20.19 / \textbf{21.12} & 30.08 / \textbf{31.18} \\ \specialrule{0em}{0.85pt}{0.85pt} 
\bottomrule[0.8pt]
\end{tabular} 
\end{center}
\caption{The many-to-many translation results on the IWSLT dataset.
Our parameter differentiation method is based on the granularity of Operation.
We compare our method with the Multilingual method and report the result with format of Multilingual/Ours.
\textbf{Bold} indicates the better result.
\label{tab:tabm2m}}
\end{table*}

\subsection{Results}

\vspace{1mm}

\textbf{OPUS}
Table \ref{tab:opus} shows the results of our method and the baseline methods on the OPUS dataset.
In both one-to-many ($\leftarrow$) and many-to-one ($\rightarrow$) directions,
our methods consistently outperform the Bilingual and Multilingual baselines
and gains improvement over the Multilingual baseline by up to $+1.40$ and $+1.55$ BLEU on average.
Compared to other parameter sharing methods,
our method achieves the best results in $20$ of $24$ translation directions
and improves the average BLEU by a large margin. 
As for the different granularities in our method,
we find that the Operation level achieves the best results on average,
due to the fine-grained control of parameter differentiation compared to the Layer level and the Module level.

For the model sizes, 
the method of \cite{DBLP:conf/wmt/SachanN18} that pre-defines the sharing modules 
increases linearly with the number of languages involved
and results in a larger model size ($3.71$x).
In our method, the model size is unrelated to the number of languages, which provides more scalability and flexibility.
Since we use different granularities instead of performing differentiation on every single parameter,
the actual sizes of our method range from $1.82$x to $2.14$x, close but not equal to the predefined $2$x.

\textbf{WMT}
We further investigate the generalization performance with experiments on the unbalanced WMT dataset.
As shown in Table \ref{tab:wmt},
the Multilingual model 
benefits lower-resource languages (ET, RO) translation
but hurts the performances of higher-resource languages (FR, CS, DE).
In contrast, 
our method gains more improvements in higher-resource language 
(+2.21 for FR$\leftarrow$EN)
than lower-resource language (+1.05 for RO$\rightarrow$EN).
Our method can also outperform the Bilingual method in $8$ of $10$ translation directions.

\textbf{IWSLT}
The results on the many-to-many translation scenario with the IWSLT dataset are shown in Table \ref{tab:tabm2m}.
Our method based on Operation level granularity outperforms the Multilingual baseline in all $20$ translation directions,
but the improvement ($+1.10$ BLEU on average) is less significant than those on the other two datasets.
The reason is that the $5$ languages in the IWSLT dataset belong to the same Indo-European language family and thus the shared parameters may be sufficient for modeling all translation directions.

\subsection{Analyses}

\paragraph{Parameter Differentiation Across Layers}

Using a shared encoder for one-to-many translation and a shared decoder for many-to-one translation 
has been proved effective and is widely used 
\cite{zoph2016multi,dong2015multitask,DBLP:conf/wmt/SachanN18}.
However, there lack of analyses on different sharing strategies across layers.
The parameter differentiation method provides a more fine-grained control of parameter sharing,
making it possible to offer such analyses.
To investigate the parameter sharing across layers,
we calculate the number of differentiation units within each layer of the final model trained with Operation level granularity.
For comparison, the completely shared model has $8$ differentiation units in each encoder layer 
and $13$ in each decoder layer (see details in supplementary  materials).

\begin{figure}[t]
\centering
\includegraphics[width=8cm]{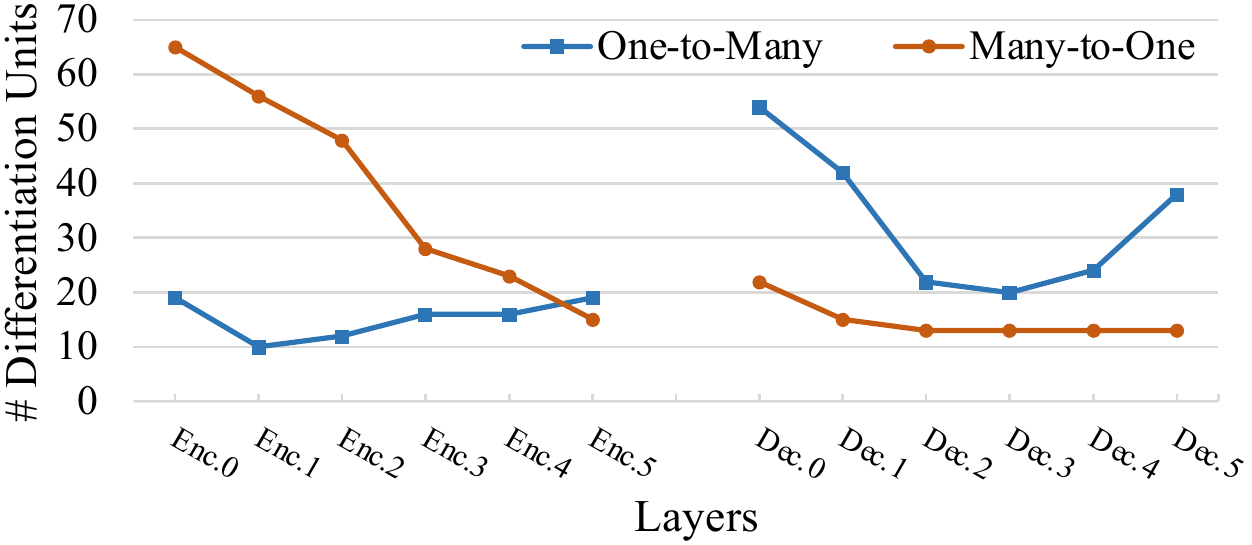}
\caption{The number of differentiation units
within each layer of the final model.\label{fig:count}}
\end{figure}

The results are shown in Figure \ref{fig:count}.
For many-to-one translation,
the task-specific parameters are mainly distributed in shallower layers of the encoder
and the parameters in the decoder tend to stay shared.
On contrary, for one-to-many translation, the decoder has more task-specific parameters than the encoder.
Different from the encoder in which shallower layers are slightly more task-specific,
both the shallower and the deeper layers are more specific than the middle layers in the decoder.
The reason is that the shallower layers in the decoder take tokens from multiple languages as input and the 
deeper layers are responsible for generating tokens in multiple languages.

\paragraph{Parameter Differentiation and Language Family}



We investigate the correlation between the parameter sharing obtained by differentiation and the language families.
Intuitively,
linguistically similar languages are more likely to have shared parameters.
To verify this, 
we first select encoder.layer-0.self-attention.value-projection,
which differentiate for the most times and is the most specialized,
and then analyze its differentiation process during training.

\begin{figure}[t]
\centering
\includegraphics[width=8cm]{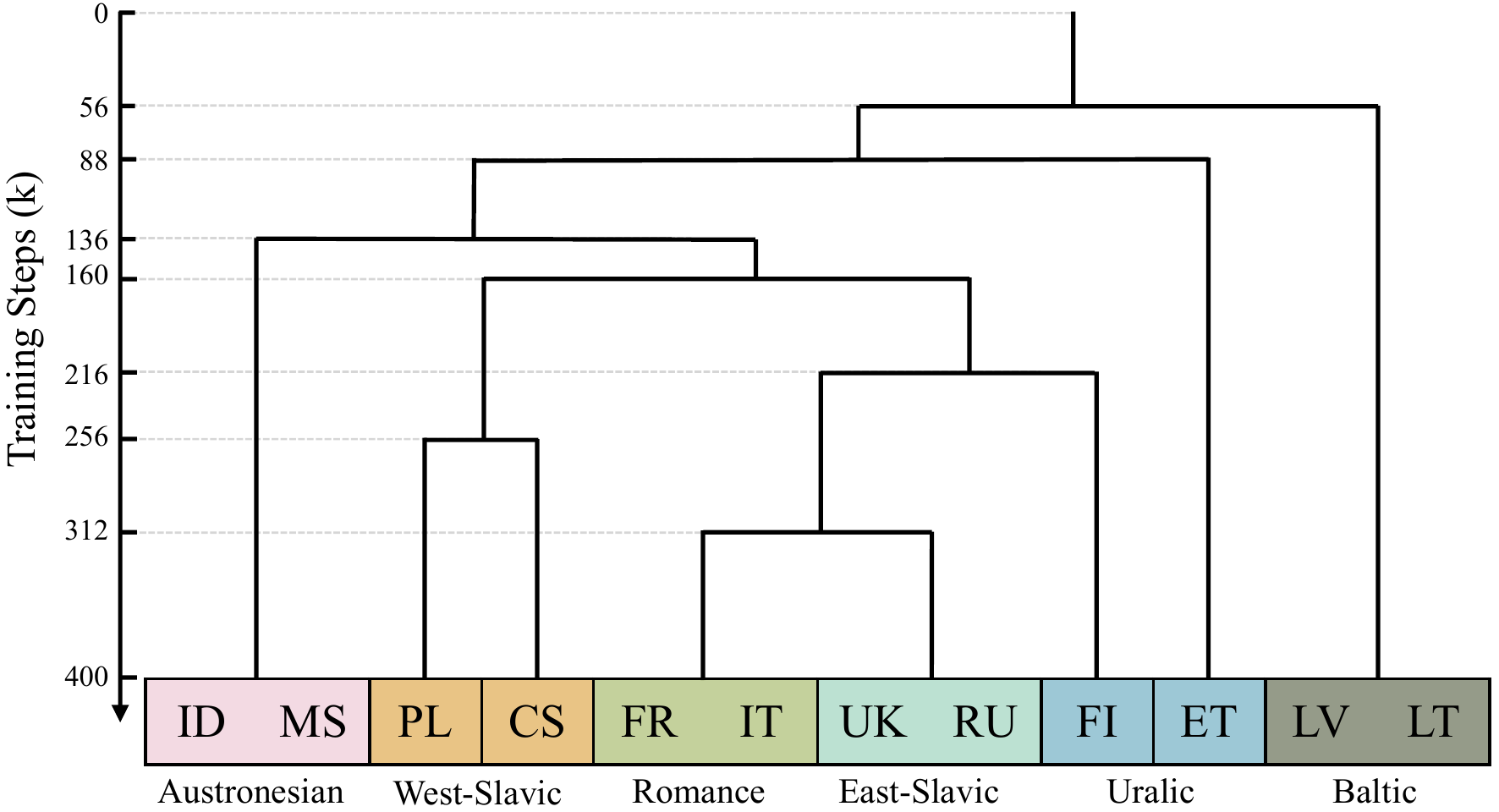}
\caption{The differentiation process of the
parameter group encoder.layer-0.self-attention.value-projection.
Parameters are shared across languages in a square and the colors represent linguistic proximities.
\label{fig:param}}
\end{figure}

Figure \ref{fig:param} shows the differentiation process of the most specialized parameter.
From the training steps, we can find that the differentiation happens aperiodically for this parameter.
As for the differentiation results,
it is obvious that the parameter sharing strategy is highly correlated with the linguistic proximity
like language family or language branch.
For example, 
ID and MS belong to the Austronesian language and share the parameters
while ID and FR belonging to the Austronesian language and the Romance language respectively
have task-specific parameters.
Another interesting observation is that the Baltic languages (LV and LT)
become specialized at the early stage of training.
We examine the OPUS dataset and find out that 
the training data of LV and LT are mainly from the political domain,
while other languages are mainly from the spoken domain.

\begin{figure}[t]
\vspace{-2mm}
\centering
\includegraphics[width=7.5cm]{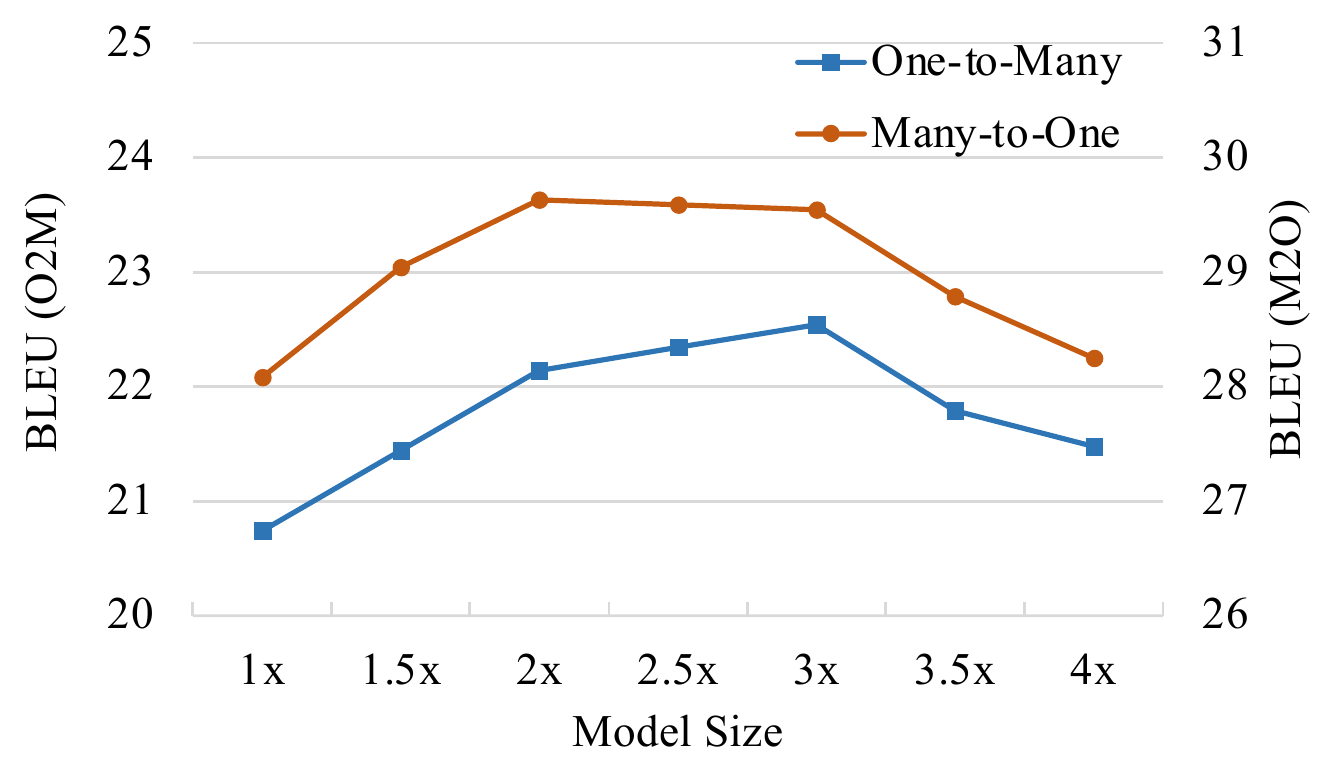}
\caption{The correlation between model size and 
the average BLEU 
over all language pairs on the OPUS dataset 
in both one-to-many and many-to-one directions.\label{fig:size}}
\end{figure}

\paragraph{The Effect of Model Size}

We notice that the model size is not completely correlated with performance
according to the results in Table \ref{tab:opus}.
Our method initialize the model as completely shared with the model size of $1$x,
and may differentiate into bilingual models in extreme cases.
The completely shared model tends to preserve the general knowledge,
while the bilingual models only capture language-specific knowledge.
To investigate the effect of the differentiation level,
we evaluate the relationship between model size and translation quality.


As shown in Figure \ref{fig:size},
the performance first increases with a higher differentiation level (larger model size)
and then decreases when the model grows over a certain threshold.
The best results are obtained with $3$x and $2$x model sizes for 
one-to-many and many-to-one directions respectively,
which indicates that 
the model needs more parameters for handling multiple target languages (one-to-many)
than multiple source languages (many-to-one).

\section{Related Work}

Multilingual neural machine translation (MNMT)
aims at handling translation between multiple languages
with a single model \cite{DBLP:journals/csur/DabreCK20mnmtsurvey}.
In the early stage,
researchers share different modules like encoder \cite{dong2015multitask},
decoder \cite{zoph2016multi}, or attention mechanism \cite{firat2016multiway}
to reduce the parameter scales in bilingual models.
The success in sharing modules motivates a more aggressive parameter sharing 
that handles all languages with a completely shared model \cite{johnson2017google,ha2016toward2}.

Despite its simplicity,
the completely shared model faces capacity bottlenecks for retaining
specific knowledge of each language \cite{aharoniJF19naacl}.
Researchers resort to language specific modeling with various parameter sharing strategies
\cite{DBLP:conf/wmt/SachanN18,wang2019compact,DBLP:conf/emnlp/WangZZXZ18three},
such as the attention module \cite{DBLP:conf/emnlp/WangZZLZ19,DBLP:conf/coling/BlackwoodBW18taskattn,DBLP:conf/aaai/HeWYZZZ21},
decoupling encoder or decoder \cite{DBLP:conf/eacl/EscolanoCFA21encdecshare},
additional adapters \cite{DBLP:conf/emnlp/BapnaF19simple},
and language clustering \cite{DBLP:conf/emnlp/TanCHXQL19cluster}.

Instead of augmenting the model with manually designed language-specific modules, 
researchers attempt to search for a language-specific sub-space of the model, such as 
generating the language-specific parameters from global ones \cite{DBLP:conf/emnlp/PlataniosSNM18contexparamgen},
language-aware model depth \cite{DBLP:conf/nips/LiSTK20latentdepth},
language-specific routing path \cite{DBLP:conf/iclr/ZhangBSF21sharenot} 
and language-specific sub-networks \cite{DBLP:conf/acl/Xie0G020,DBLP:conf/acl/LinWWL20langsubnet}.
These methods start from a large model that covers all translation directions,
where the size and structure of the initial model are non-trivial to determine.
While our method initializes a simple shared model
and lets the model to automatically grows into a more complicated one,
which provides more scalability and flexibility.

\section{Conclusion and Future Work}
In this paper,
we propose a novel parameter differentiation based method that can 
automatically determine which parameters should be shared 
and which ones should be language-specific.
The shared parameters can dynamically differentiate into more specialized types during training.
The extensive experiments on three multilingual machine translation datasets verify the effectiveness of our method.
The analyses reveal that the parameter sharing configurations obtained by our method are highly correlated with 
the linguistic proximities.
In the future, we want to let the model learn when to stop differentiation
and explore other differentiation criteria for more multilingual scenarios 
like the zero-shot translation and the incremental multilingual translation.

\section{Acknowledgement}
The research work descried in this paper has been supported by the Natural Science Foundation of China under Grant No. U1836221 and 62122088, 
and also by Beijing Academy of Artificial Intelligence (BAAI).
\bibliography{aaai22}

\end{document}